\documentclass[letterpaper, 10 pt, conference]{ieeeconf}  % Comment this line out if you need a4paper

\IEEEoverridecommandlockouts                              % This command is only needed if 
\usepackage{graphicx} % for pdf, bitmapped graphics files
\usepackage{multirow}
\usepackage{times} % assumes new font selection scheme installed
\usepackage{amsmath} % assumes amsmath package installed
\usepackage{amssymb}  % assumes amsmath package installed
\usepackage{algorithm}
\usepackage{algorithmic}
\usepackage{booktabs}
\usepackage{euscript}
\usepackage{pgfplots}
\usepackage{tikz}

\DeclareMathOperator{\x}{\mathbf{x}}
\DeclareMathOperator{\tensor}{T^{\frac{1}{2}}}

\title{\LARGE \bf
Real-Time Variational Fisheye Stereo \\without Rectification and Undistortion
}

\author{Menandro Roxas$^{1}$ and Takeshi Oishi$^{2}$% <-this % stops a space
\thanks{Both authors are with The University of Tokyo,
        Tokyo, Japan
        {$^{1}$\tt\small roxas, $^{2}$\tt\small oishi@cvl.iis.u-tokyo.ac.jp}}%
}

\begin{document}

\maketitle
\thispagestyle{empty}
\pagestyle{empty}

%%%%%%%%%%%%%%%%%%%%%%%%%%%%%%%%%%%%%%%%%%%%%%%%%%%%%%%%%%%%%%%%%%%%%%%%%%%%%%%%
\begin{abstract}
Dense 3D maps from wide-angle cameras is beneficial to robotics applications such as navigation and autonomous driving. In this work, we propose a real-time dense 3D mapping method for fisheye cameras without explicit rectification and undistortion. We extend the conventional variational stereo method by constraining the correspondence search along the epipolar curve using a trajectory field induced by camera motion. We also propose a fast way of generating the trajectory field without increasing the processing time compared to conventional rectified methods. With our implementation, we were able to achieve real-time processing using modern GPUs. Our results show the advantages of our non-rectified dense mapping approach compared to rectified variational methods and non-rectified discrete stereo matching methods.

\end{abstract}

%%%%%%%%%%%%%%%%%%%%%%%%%%%%%%%%%%%%%%%%%%%%%%%%%%%%%%%%%%%%%%%%%%%%%%%%%%%%%%%%
\section{INTRODUCTION}
Wide-angle (fisheye) cameras have seen significant usage in robotics applications. Because of the wider field-of-view (FOV) compared to pinhole camera model, fisheye cameras pack more information in the same sensor area which are advantageous especially for object detection, visual odometry, and 3D reconstruction.

Real-time dense 3D mapping using fisheye cameras have several advantages especially in navigation and autonomous driving. For example, the wide field-of-view allows simultaneous visualization and observation of objects in different directions.

Several methods have addressed the 3D mapping problem for fisheye cameras. The most common approach performs rectification of the images to perspective projection which essentially removes the main advantage of such cameras - wide FOV. Moreover, information closer to the edge of the image are highly distorted while objects close the center are highly compressed, not to mention adding unnecessary degradation of image quality due to spatial sampling. Other rectification that retains the fisheye's wide FOV involves reprojection on a sphere, which suffers from similar degradation especially around the poles. We address these issues by directly processing the distorted images without rectification or undistortion.

We embed our method in a variational framework, which inherently produces smooth dense maps in contrast to discrete stereo matching methods. We propose to use a trajectory field that constrains the search space of corresponding pixels along the epipolar curve. We also propose a fast way of generating the trajectory field that does not require additional processing time compared to conventional variational methods.

The advantage of our proposed method is two-folds. First, without rectification or undistortion, the sensor level image quality is preserved. Second, our method can handle arbitrary camera distortions. While the results in this paper focuses on fisheye cameras, applying our method on other camera models is straightforward.

Our results show additional accurate measurements when compared to conventional rectified methods, and more accurate and dense estimation compared to non-rectified discrete methods. Finally, with our implementation, we were able to achieve real-time processing on a consumer fisheye stereo camera system and modern GPUs.

%Disadvantage of rectification methods and advantage of variational methods. Directly processing distorted images reduces additional degradation of the image due to sampling of the images.

%In classical stereo depth estimation BLAH BLAH the BLAH BLAH along the epipolar lines BLAH BLAH. With omnidirectional cameras, the epipolar lines between two views are BLAH BLAH as curves CITE. Without rectifying the images, we can estimate the disparity by constraining the correspondence along the epipolar curves. 

%In this work, we will BLAH BLAH a sweeping generalization of different camera models as omnidirectional cameras. However, panoramic systems are developed WARPED and STITCHED version of the BLAH BLAH. On the sensor level, consumer cameras that are capable of wide angle views are mostly fisheye cameras, excluding catadioptric systems which are usually employed in laboratories. Hence, we will limit our results on fisheye camera systems, but our proposed method is applicable to all omindirectional cameras.

\begin{figure}
	%\fbox{\rule{0pt}{1.5in} \rule{0.9\linewidth}{0pt}}
	\centering
	\begin{tabular}{cc}
		\includegraphics[width=0.22\textwidth]{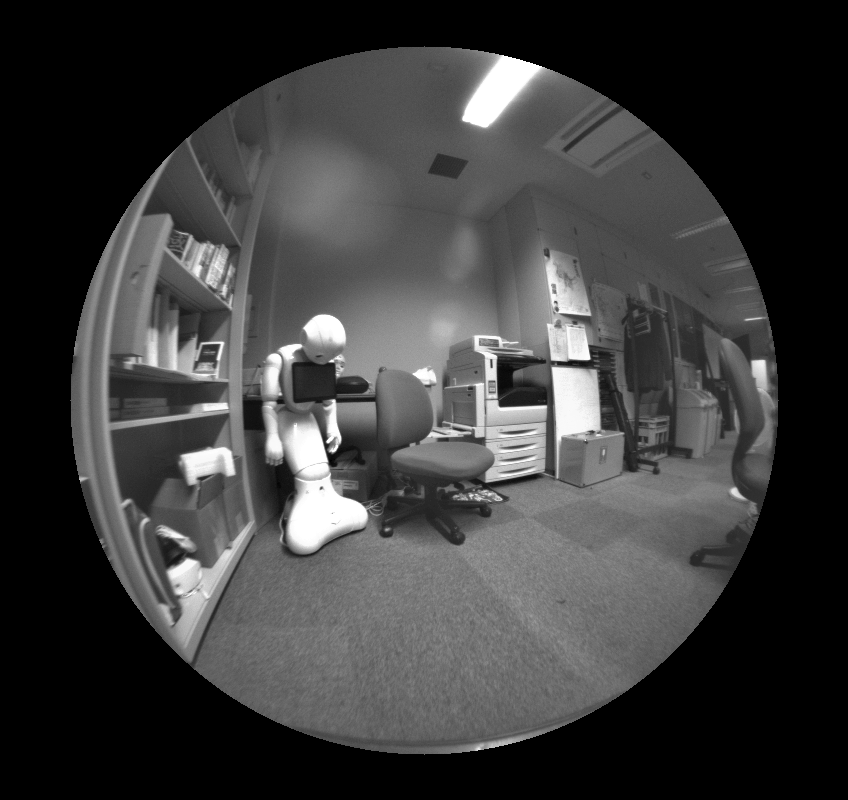} &%
		\includegraphics[width=0.22\textwidth]{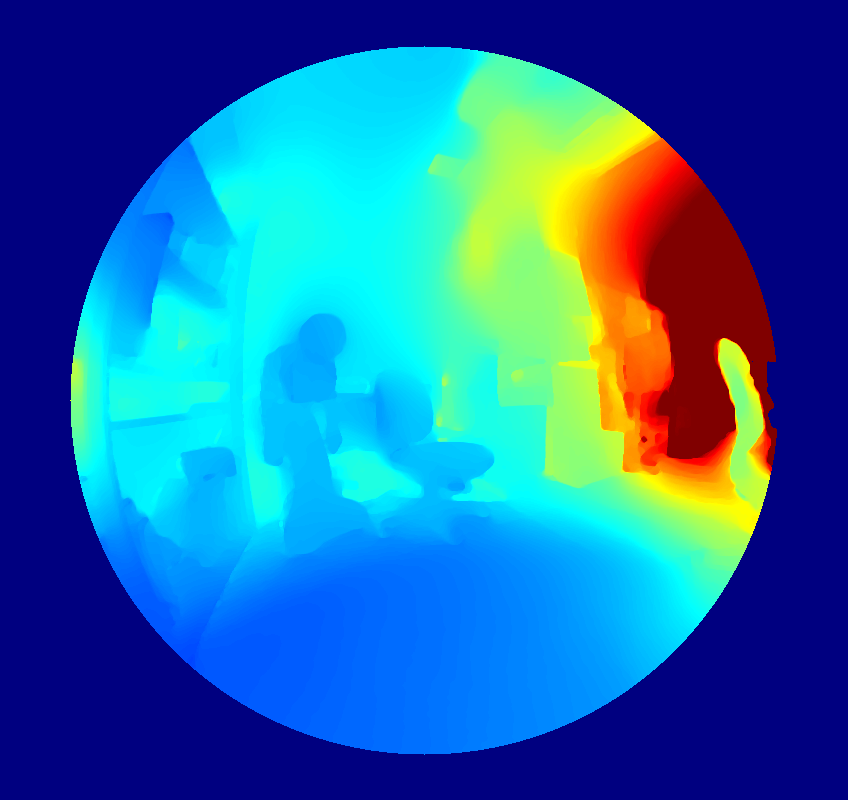}
	\end{tabular}
	\caption{Non-rectified variational stereo method result on a fisheye stereo camera.}
	\label{fig:stereocamera}
\end{figure}

%\begin{figure}
%	\fbox{\rule{0pt}{1.5in} \rule{0.9\linewidth}{0pt}}
%	\caption{Sample results.}
%	\label{fig:overviewresults}
%\end{figure}

\section{RELATED WORK}
Dense stereo estimation in perspective projection consists of a one-dimensional correspondence search along the epipolar lines. In a variational framework, the search is akin to linearizing the brightness constancy constraint along the epipolar lines. In \cite{stuhmer2010}, a differential vector field induced by arbitrary camera motion was used for linearization. However, their method, as with other variational stereo methods in perspective projection such as \cite{ranftl2012}, requires undistortion and/or rectification (in case of binocular stereo) to be applicable for fisheye cameras \cite{schneider2016}.

Instead of perspective rectification, some method reprojects the images to spherical or equirectangular projection \cite{arican2007} \cite{schonbein2014} \cite{bagnato2011} \cite{gao2017}. However, this approach suffers greatly from highly distorted images along the poles which makes estimation less accurate especially when using the variational framework. Similar to our brightness constancy linearization approach, \cite{bagnato2011} generates differential vectors induced by variations on a 2-sphere in which the variational stereo method was applied. However, their graph-based formulation is a solution to the self-induced problem arising from reprojecting the image on a spherical surface. In contrast, our method does not require reprojection on a 2-sphere and works directly on the distorted images without undistortion, reprojection or rectification. We do this by evaluating the variations directly from the epipolar curve.

%In , fisheye cameras were used Similar to fisheye cameras, catadioptric cameras have been used for dense stereo estimation\cite{schonbein2014} . However, these methods rectifies the image to spherical or equirectangular projection to simplify the search along a line.  Other methods include rectifying the omnidirectional view to four-sided \cite{arican2007} 
%\cite{schneider2016} \cite{gao2017} \cite{schonbein2014} \cite{arican2007} \textit{Nunc non magna in lacus porttitor sagittis at sit amet nisi. Fusce tincidunt lacus dolor. Duis dignissim, sem non commodo auctor, arcu libero ullamcorper tellus, eget posuere nunc velit quis mauris. Mauris blandit sapien quis tortor consequat ornare. Vestibulum ante ipsum primis in faucibus orci luctus et ultrices posuere cubilia Curae; Mauris sit amet odio vulputate, vestibulum ante luctus, faucibus mi. Sed sollicitudin massa at dapibus iaculis. Aenean eget bibendum dui, at finibus ligula. Aenean sit amet porta velit. Praesent commodo metus ut mi imperdiet interdum. }

Other methods also directly work on the distorted fisheye images. In \cite{lsdslam2015}, the unified camera model \cite{unifiedcameramodel} was used to determine the path of the search space, which are incrementally shifted (akin to differential vectors) from a reference pixel to the maximum disparity. At each point, the projection function is re-evaluated which the authors claim was costly compared to linear search. Their mapping method, while real-time, only produces semi-dense depth maps. In \cite{bunschoten2003}, a similar parameterization of the epipolar curve was done, but only applied on window-based stereo matching. Other methods adapts linear matching algorithms to omni-directional cameras such as semi-global matching \cite{khomutenko2016}, plane-sweeping \cite{fisheyeplanesweep2014} and a variant called sphere-sweeping \cite{inso2016}. Sparse methods were also adapted to handle fisheye distortion such as \cite{sfmfisheye} among others. 
%was extended to the unified camera models and the epipolar curve was traced through rasterization \cite{rasterization1990}. In , the plane-sweeping algorithm was adapted to fisheye camera models. On the other hand, a variant of the plane-sweeping algorithm, sphere sweeping, was proposed in \cite{inso2016}.

%However, this can be addressed by solving the variations from the epipolar curve of the distorted images. 
%is highly discretized which restricts the differential vectors to be defined in BLAH BLAH where in our epipolar curve approach BLAH BLAH. Moreover,  suffer from unnecessary BLAH BLAH when projected on the spherical surface. 

\section{VARIATIONAL FISHEYE STEREO}
In this section, we will first introduce the problem of image linearization for fisheye cameras in Sec. \ref{sec:dataterm}. We will then propose our trajectory field and warping techniques in Secs. \ref{sec:trajectoryfield} and \ref{sec:warping}. Finally, we will combine our approach with a variational optimization method and summarize the algorithm in Sec. \ref{sec:algorithm}.

\subsection{Image Constancy Constraint}
\label{sec:dataterm}
Classical variational stereo method consists of solving a dense disparity map between a pair of images that minimizes a convex energy function. Given two images, $I_0$ and $I_1$, with known camera transformation and intrinsic parameters, the one-dimensional disparity $u$ can be solved by minimizing:
\begin{equation}
E(u) = E_{data}(u) + E_{smooth}(u)%\left|\rho(\x, u)\right|d^2\x + \phi(u)d^2\x
\label{eq:energyterm}
\end{equation}
The above functional consists of a data term and a smoothness term. Building upon perspective camera stereo methods \cite{stuhmer2010}\cite{ranftl2012}, we only need to modify the formulation of the data term in order to accommodate the distortion effects in fisheye cameras. 

In general, the data term penalizes the residual, $\rho$, which measures the constancy between corresponding pixels in $I_0$ and $I_1$. $I$ can be any value such as brightness, intensity gradient, non-local transforms \cite{censustransform}, etc. For fisheye cameras, these correspondences are constrained along the epipolar curve, $\gamma: \mathbb{R} \to \mathbb{R}^2$ and finding them constitutes a one-dimensional search \cite{lsdslam2015}\cite{khomutenko2016}\cite{fisheyeplanesweep2014} along $\gamma$. In our case, we will solve the correspondences using a variational framework. 

Let $\x$ be a point on the epipolar curve. The corresponding point at a distance $u$ along the curve can be expressed as $\pi(\exp(\hat{\xi}_1)\cdot \mathbf{X}(\x, u))$, where $\pi : \mathbb{R}^3 \to \mathbb{R}^2$ is the projection of the 3D point $\mathbf{X}$ on the image plane $\Omega_1$ of $I_1$, and $\hat{\xi}_1$ is the camera pose of $I_1$ relative to $I_0$. We can then express the residual as:
\begin{equation}
\rho(\x, u) =  I_1\left(\pi\left(\exp(\hat{\xi}_1) \cdot \mathbf{X}(\x, u)\right)\right) - I_0\left(\x\right)
\label{eq:dataterm}
\end{equation}
Assuming that $I_0$ and $I_1$ is linear along the curve, we can approximate (\ref{eq:dataterm}) using the first-order Taylor expansion, and using a simplified notation $I_1(\x,u)$ as in \cite{stuhmer2010}
\begin{equation}
\bar{\rho}(\x, u) = I_1(\x, u_{\omega}) + \left.(u-u_{\omega})\frac{d}{du}I_1(\x, u) \right\vert_{u_{\omega}} - I_0(\x)
\label{eq:linearization}
\end{equation}
So far, our formulation of the data term still follows that of \cite{stuhmer2010}. 

Formally, the derivative $\frac{d}{du}I_1(\x,u)$ can be expressed as the dot product of the gradient of $I_1(\x, u)$ and a differential vector at $\x$:
\begin{equation}
\frac{d}{du}I_1(\x,u) = \nabla I_1(\x, u) \cdot \underbrace{\frac{d}{du}\pi\left(\exp(\hat{\xi}_1)\cdot \mathbf{X}(\x, u)\right)}_\text{differential vector}
\label{eq:differentialvector}
\end{equation}
However, in practice, we directly solve for the variations of $I$ along the epipolar curve. In discrete form, we have:
\begin{equation}
\frac{d}{du}I_1(\x,u) = I_1(\x + \gamma') - I_1(\x)
\label{eq:variations}
\end{equation}
where $\gamma'$ is the differential vector. (Note that Eqs. (\ref{eq:differentialvector}) and (\ref{eq:variations}) result in the pole-stretching problem when using spherical projections \cite{bagnato2011}).

Minimizing Eq. (\ref{eq:linearization}) results in the incremental disparity $(u-u_{\omega})$ which we will designate from here on as $\delta u_{\omega}$. For small $\delta u_{\omega}$, the differential vector in (\ref{eq:differentialvector}) is equal the tangential vector of the epipolar curve $\gamma' = \nabla \gamma$. 

Moreover, since the linearity assumption of $I$ is only valid for a small disparity, (\ref{eq:linearization}) is usually embedded in an iterative warping framework \cite{warping} around a known disparity $u_{\omega}$ (hence, the term $I_1(\x, u_{\omega})$). That is, for every warping iteration $\omega$, we update $u_{\omega + 1} = u_{\omega} + \delta u_{\omega}$.
%\begin{equation}
%u_{\omega + 1}  = u_{\omega} + u
%\label{eq:warping}
%\end{equation}
%%BLAH BLAH (\ref{eq:warping}) is iterated BLAH BLAH for every STEP BLAH BLAH.

%I THINK I NEED TO EXPLAIN THE WARPING HERE

This formulation raises two issues when used in a fisheye camera system. 
\begin{itemize}
	\item First, the warping technique requires a re-evaluation of $\gamma$ at every iteration to find the tangential vectors at $u_{\omega}$ which is tedious and time consuming. 
	\item Second, even if we assume that the image is perfectly linear along the epipolar curve, $\nabla I$ is only valid along the direction of the tangential vectors. In a perspective projection, this is not a problem since the tangential vectors indicates the exact direction of the epipolar lines. In our case, the gradient will need to be evaluated exactly along the curve. 
\end{itemize}

In the following sections, we will elaborate on our approach to solve these two issues. % in the following sections. %Fortunately, this can be addressed by limiting the value of $u_{\omega}$. 
\begin{figure}
	%\fbox{\rule{0pt}{1.5in} \rule{1.0\linewidth}{0pt}}
	\includegraphics[width=1.0\linewidth]{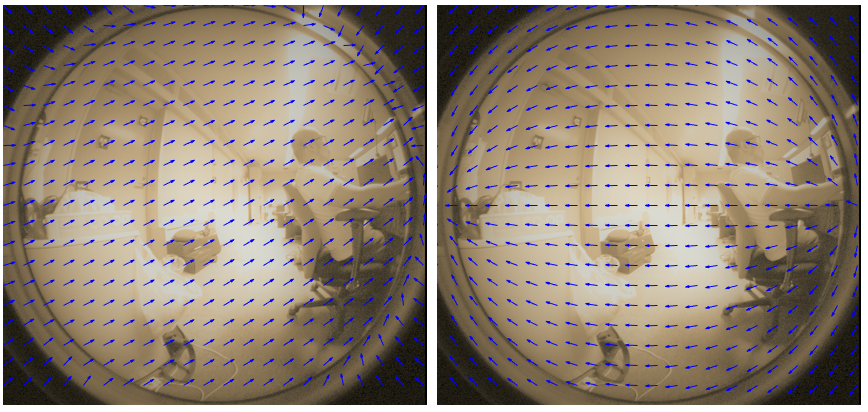}
	\caption{Calibration (left) and trajectory (right) field for a binocular fisheye stereo.}
	\label{fig:trajectoryfield}
\end{figure}

\subsection{Trajectory Field Representation for Epipolar Curves}
\label{sec:trajectoryfield}
To avoid the re-evaluation of the epipolar curve for every warping step, we generate a trajectory field image that represents the tangential vectors $\gamma'$ at every pixel $\x$. As a result, $\gamma'$ at the next iteration step can be simply solved using bicubic interpolation.

Moreover, instead of solving for the parameterized curve function for every pixel \cite{svoboda1998}, we programmatically generate the trajectory field. We first assume a known transformation $\hat{\xi}_1$ between two camera positions with non-zero translation ($\left|t\right| \neq 0$) and known projection $\pi$. Our method is not restricted on any type of camera model \cite{unifiedcameramodel} \cite{enhanced} \cite{kannala2006} as long as the projection function $\pi$ is defined. 

Using $\pi$, we project a surface of arbitrary depth onto the two cameras: $\x_0 = \pi(\mathbf{X})$,  $\x_1 = \pi(\exp(\hat{\xi}_1)\cdot \mathbf{X})$. This gives us the exact correspondence $\mathbf{w}(\x_0,\x_1) = \x_1 - \x_0$. In a perspective projection, this mapping or the optical flow already represents the slope of the epipolar lines. Assuming pre-rotated images, i.e. $R=0$, the direction of the optical flow, $\frac{\mathbf{w}}{\left|\mathbf{w}\right|}$, will be dependent only on the direction of the camera translation $t$ and independent of $\left|t\right|$ and the surface depth $\left|\mathbf{X}\right|$. However, for fisheye projection, $\frac{\mathbf{w}}{\left|\mathbf{w}\right|}$ is still affected by the camera distortion. 

To address this, we can represent the optical flow as the sum of the tangential vectors along the path of the epipolar curve between the two corresponding points. Let the parameterization variable for $\gamma$ be $s = [0,1]$. In continuous form, we can express $\mathbf{w}(\x_0,\x_1)$ as:
\begin{equation}
\mathbf{w}(\x_0,\x_1) = \left.\int_0^c \gamma'(s)ds \right|_{c=1}%\sum_{i=0}^{n} \gamma'(\x_i)
\label{eq:sumangle}
\end{equation}
By scaling the camera translation such that $\left|t\right| \to 0$, the left hand side of (\ref{eq:sumangle}) approaches $\mathbf{0}$. It follows that the right hand side becomes:
\begin{equation}
\label{eq:limit}
\lim_{c \to 0} \int_0^c \gamma'(s)ds = \gamma'(0)
\end{equation}
which finally allows us to approximate $\gamma'(0) \approx \frac{\mathbf{w}}{\left|\mathbf{w}\right|}$. In short, $\frac{\mathbf{w}}{\left|\mathbf{w}\right|}$ gives us the normalized trajectory field.

\subsection{Warping Technique}
\label{sec:warping}
\begin{figure}
	\begin{center}
		\includegraphics[width=0.6\linewidth]{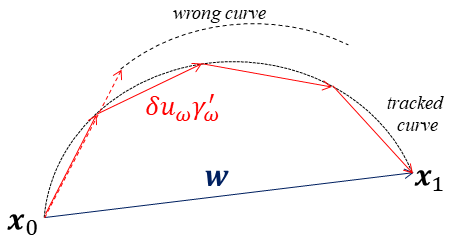}
		\caption{Epipolar curve as a piecewise linear function. Large incremental $\delta u_{\omega}$ results in wrong tracked curve.}
		\label{fig:piecewiselinear}
	\end{center}
	%\fbox{\rule{0pt}{1.5in} \rule{1.0\linewidth}{0pt}}
\end{figure}

The trajectory field discretizes the epipolar curve by assigning finite vector values for every pixel. We can think of this approach as decomposing the epipolar curve as a piecewise linear function (see Figure \ref{fig:piecewiselinear}) which allows us to express the disparity $u$ as:
\begin{equation}
u = \sum_{{\omega}=0}^{N}\delta u_{\omega}
\label{eq:piecewise}
\end{equation}
%where $\delta u_i$ are the incremental solutions to (\ref{eq:linearization}). 
where $N$ is the total number of warping iterations. 

Clearly, we can better approximate the epipolar curve by setting a magnitude limit to the incremental $\delta u_{\omega}$ and increasing the number of iterations $N$. Moreover, doing so also prevents missing the correct trajectory of the curve since $\delta u_{\omega}$ is constrained along $\gamma'_{\omega}$. (see Figure \ref{fig:piecewiselinear}).

To solve the final warping vector of $\x$ using the trajectory field, we use:
\begin{equation}
\mathbf{w} = \sum_{{\omega}=0}^{N}\delta u_{\omega} \gamma'_{\omega}
\label{eq:warpingvector}
\end{equation}

%To address the second issue raised in Section \ref{sec:dataterm}, BLAH BALH limit the value of $u_{\omega}$ after every iteration.

\subsection{Anisotropic TGV-L1 Optimization}
\label{sec:algorithm}
Before we continue, we will complete the energy functional in (\ref{eq:energyterm}). We will follow the anisotropic tensor-guided total generalized variation (TGV) constraint described in \cite{ranftl2012} and combine it with data term (\ref{eq:dataterm}) which results in the following:
\begin{align}
\label{eq:tgvl1}
E(u) = & \lambda\int_{\Omega}\left|\rho(\x, u)\right|d^2\x + \nonumber \\ 
& \alpha_0 \int_{\Omega}\left|\nabla v\right|d^2\x + 
\alpha_1 \int_{\Omega}\left|T^{\frac{1}{2}}\nabla u - v\right|d^2\x
\end{align}

We can minimize (\ref{eq:tgvl1}) using primal-dual algorithm, which consists of a gradient-ascent on the dual variables $p: \mathbb{R}^2$ and $q: \mathbb{R}^4$, followed by a gradient-descent and over-relaxation refinement step on the primal variables $u$ and $v: \mathbb{R}^2$. The algorithm is summarized as:
\begin{equation}
\label{eq:primaldual}
\begin{cases}
	&p_{k+1} = \mathcal{P}\left(p_k + \sigma_p \alpha_1 (\tensor \nabla \bar{u}_k - \bar{v}_k) \right)  \\
	&q_{k+1} = \mathcal{P}\left( q_k + \sigma_q \alpha_0 (\nabla \bar{v}_k)\right)  \\
	&u_{k+1} = (I + \tau_u \partial G)^{-1}(u_n + \tau_u div(\tensor p_{k+1})) \\
	&v_{k+1} = v_k + \tau_v (div q_{k+1} + p_{k+1}) \\
	&\bar{u}_{k+1} = u_{k+1} + \theta(u_{k+1} - \bar{u}_k) \\
	&\bar{v}_{k+1} = v_{k+1} + \theta(v_{k+1} - \bar{v}_k) \\
\end{cases}
\end{equation}
where $\mathcal{P}(\phi) = \frac{\phi}{max(1, \|\phi\|)}$ is a fixed-point projection operator. The step sizes $\tau_u>0, \tau_v>0, \sigma_u>0, \sigma_v>0$ are solved using a pre-conditioning scheme following \cite{preconditioning} while the relaxation variable $\theta$ is updated for every iteration as in \cite{firstorderapprox}. The tensor $\tensor$ is calculated as:
\begin{equation}
\label{eq:tensor}
\tensor = \text{exp}(-\beta \left|I_0\right|^{\eta})nn^T + n^{\bot} n^{\bot T}
\end{equation}
where $n = \frac{\nabla I_0}{\left|\nabla I_0\right|}$ and $n^{\bot}$ is the vector normal to $\nabla I_0$, while $\beta$ and $\eta$ are scalars controlling the magnitude and sharpness of the tensor. This tensor guides the propagation of the disparity information among neghboring pixels, while considering the natural image boundaries as encoded in $n$ and $n^{\bot}$.

The resolvend operator \cite{firstorderapprox} $(I + \tau_u \partial G)^{-1}(\hat{u})$ is evaluated using the thresholding scheme:
\begin{equation}
\label{eq:thresholding}
(I + \tau_u \partial G)^{-1}(\hat{u}) = \hat{u} + 
\begin{cases}
\tau_u \lambda I_u &\text{if} \, \bar{\rho} < -\tau_u \lambda I_u^2 \nonumber \\
-\tau_u \lambda I_u &\text{if} \, \bar{\rho} > \tau_u \lambda I_u^2 \nonumber \\
\bar{\rho}/I_u & \text{if} \left|\bar{\rho}\right| \leq \tau_u \lambda I_u^2
\end{cases}
\end{equation}
where $I_u = \frac{d}{du}I_1(\x,u)$. We summarize our approach in Algorithm \ref{algo:main}.

\begin{algorithm}[b]
	\begin{algorithmic}
		\REQUIRE{$I_0$, $I_1$, $\hat{\xi}_1$, $\pi$}
		\STATE{Generate trajectory field} (Sec. \ref{sec:trajectoryfield})
		\STATE{$\omega=0$, $\mathbf{w_{\omega}}=0$, $u_{\omega}=0$}
		\WHILE{$\omega < N$}
			\STATE{Warp $I_1$ using $\mathbf{w_{\omega}}$}
			\WHILE{$k < nIters$}
				\STATE{Update primal-dual variables (\ref{eq:primaldual})}
			\ENDWHILE
			\STATE{Clip $\delta u_{\omega}$ (Sec. \ref{sec:warping})}
			\STATE{$u_{\omega+1} = u_{\omega} + \delta u_{\omega}$}
			\STATE{$\mathbf{w_{\omega+1}}=\mathbf{w_{\omega}} + \delta u_{\omega}\gamma'(\mathbf{x})$}
		\ENDWHILE
	\end{algorithmic}
	\caption{Algorithm for anisotropic TGV-L1 stereo for fisheye cameras.}
	\label{algo:main}
\end{algorithm}

The solved disparity is converted to depth by triangulating the unprojection rays using the unprojection function $\pi^{-1}$. This step is specific for the camera model used, hence we will not elaborate on methods to address this. Nevertheless, some camera models have closed-form unprojection function \cite{unifiedcameramodel} \cite{enhanced} while others require non-linear optimizations \cite{kannala2006}.

%\subsection{Optimization and Algorithm}
%Discuss how warping technique fits well into the proposed method.
%Range limiting (works)

\section{IMPLEMENTATION}
In this section, we discuss our implementation choices to achieve accurate results and real-time processing, which includes image pre-processing, large displacement handling and our selected optimization parameters and hardware considerations.

\subsection{Pre-rotation and calibration}
We perform a calibration and pre-rotation of the image pairs before running the stereo estimation. We create a calibration field in the same manner as the trajectory field. The calibration field contains the rotation information as well as the difference in camera intrinsic properties (for binocular stereo case). 

Again, we project a surface of arbitrary depth on the two cameras with projection funtion $\pi_0$ and $\pi_1$ while setting the translation vector $t=\mathbf{0}$. We then solve for the optical flow $\mathbf{w} = \x_1 - \x_0$. In this case the optical flow exactly represents the calibration field (see Figure \ref{fig:trajectoryfield}). In case where $\pi_0 \neq \pi_1$, such as in binocular stereo, the calibration field will also contain the difference in intrinsic properties. For example, a difference in image center results in the diagonal warping in our binocular camera system as seen in Figure \ref{fig:trajectoryfield}. Using the calibration field, we warp the second image $I_1$ once, resulting in a translation only transformation.

\subsection{Coarse-to-Fine Approach}
Similar to most variational framework, we employ a coarse-to-fine (pyramid) technique to handle large displacement. Starting from a coarser level of the pyramid, we run $N$ warping iterations and upscale both the current disparity and the warping vectors and carry the values on to the finer level. 

One caveat of this approach on a fisheye image is the boundary condition especially for gradient and divergence calculations. To address this, we employ the Neumann and Dirichlet boundary conditions applied on a circular mask that rejects pixels greater than the desired FOV. The mask is scaled accordingly using nearest-neighbor interpolation for every level of the pyramid. Moreover, by applying a mask, we also avoid the problem of texture interpolation with a zero-value during upscaling when the sample falls along the boundary of the fisheye image.

\subsection{Timing Considerations}
We implemented our method with C++/CUDA on an i7-4770 CPU and NVIDIA GTX 1080Ti GPU. For TGV-L1 optimization and primal-dual algorithm, we use the parameter values: $\beta=9.0$, $\eta=0.85$, $\alpha_0=17.0$ and $\alpha_1=1.2$. Moreover, we fix the iteration values based on the desired timing and input image size. For an 800x800 image, we found that the primal-dual iteration of $10$ is sufficient for our application, with pyramid size $=5$ and scaling $=2.0$ (minimum image width = $50$). 

For the warping iteration, we plot the trade-off between accuracy and processing time in Figure \ref{fig:timing} with fixed $\delta u^{max} = 0.2px$. From the plot, we can see that the timing linearly increases with the number of iterations, but the accuracy exponentially decreases. Choosing a proper value for $N$ needs careful considerations according to the application.
\pgfplotsset{width=7cm,compat=1.3}
\begin{figure}
	\begin{center}
		\begin{tikzpicture}
		\pgfplotsset{
			xmin=1, xmax=20
		}
		
		\begin{axis}[
		axis y line*=left,
		ymin=0, ymax=70,
		xlabel=warping iteration(N),
		ylabel=\textcolor{magenta}{\% erroneous pixel},
		font=\scriptsize,
		width=0.6\linewidth,
		height=2in,
		]
		\addplot[color=magenta, mark=o]
		coordinates{
			(1,36.40)(2,32.07)(3,27.68)(4,25.08)(5,23.52)(6,22.06)(7,20.81)(8,19.63)(9,18.63)(10,17.61)(11,16.75)(12,15.99)(13,15.17)(14,14.44)(15,13.74)(16,13.12)(17,12.56)(18,12.09)(19,11.53)(20,11.11)
		};\label{plot_one}
		%\addlegendentry{$\tau>5$px}
		\addplot[color=magenta, mark=x]
		coordinates{
			(1,51.29)(2,45.78)(3,42.97)(4,40.94)(5,38.96)(6,36.58)(7,34.73)(8,32.95)(9,31.33)(10,29.70)(11,28.69)(12,27.81)(13,26.90)(14,26.05)(15,25.21)(16,24.56)(17,23.95)(18,23.34)(19,22.73)(20,22.34)
		};\label{plot_two}
		%\addlegendentry{$\tau>1$px}
		\end{axis}
		
		\begin{axis}[
		axis y line*=right,
		axis x line=none,
		ymin=0, ymax=200,
		ylabel=\textcolor{blue}{time (ms)},
		legend pos=north west,
		font=\scriptsize,
		width=0.6\linewidth,
		height=2in,
		legend style={nodes={scale=0.8, transform shape}},
		]
		\addlegendimage{/pgfplots/refstyle=plot_one}\addlegendentry{$\tau>5$px}
		\addlegendimage{/pgfplots/refstyle=plot_two}\addlegendentry{$\tau>1$px}
		\addplot[color=blue, mark=triangle]
		coordinates{
			(1,13.00)(2,22.00)(3,33.00)(4,42.00)(5,52.00)(6,62.00)(7,71.00)(8,81.00)(9,89.00)(10,100.00)(11,106.00)(12,120.00)(13,129.00)(14,136.00)(15,144.00)(16,156.00)(17,164.00)(18,175.00)(19,185.00)(20,193.00)
		}; \addlegendentry{timing}
		\end{axis}
		
		\end{tikzpicture}
		%\fbox{\rule{0pt}{1.5in} \rule{0.9\linewidth}{0pt}}
		%\includegraphics[width=0.6\linewidth]{images/plottiming2.png}
	\end{center}
	\caption{Trade-off between accuracy and processing time for choosing the warping iteration (better viewed in color)}
	\label{fig:timing}
\end{figure}
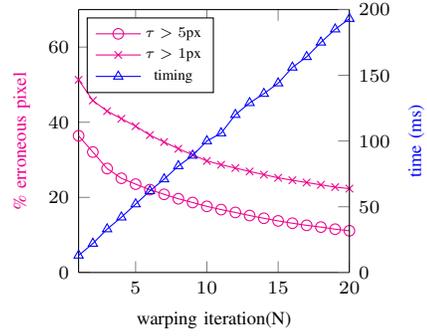

\begin{figure*}
	\centering
	\begin{minipage}{0.70\linewidth}
		\centering
		%\fbox{\rule{0pt}{1.5in} \rule{0.9\linewidth}{0pt}}
		\includegraphics[width=1.0\linewidth]{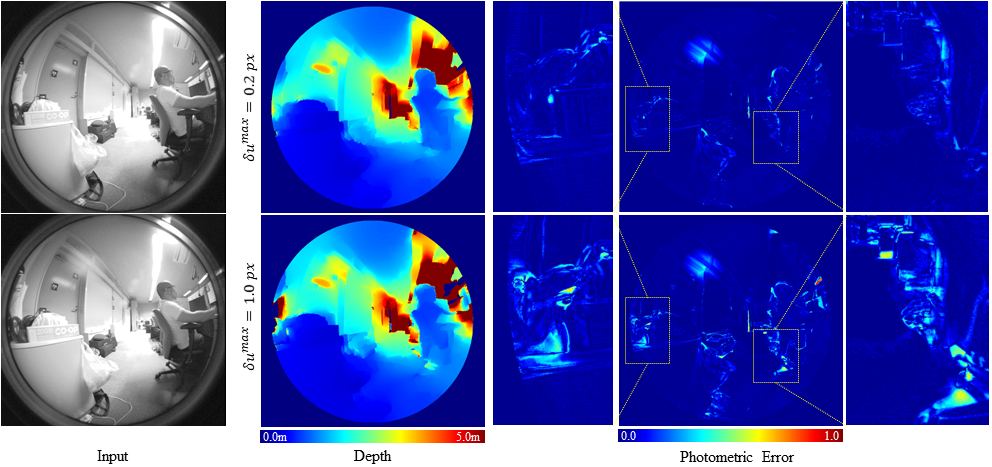}
		\caption{Limiting the magnitude of $\delta u$ per iteration reduces the error around sharp image gradients and occlusion boundaries.}
		\label{fig:limitingrange}
	\end{minipage}\hfill
	\begin{minipage}{0.27\linewidth}
		\centering
		%\resizebox{\linewidth}{1.5in}{%
		\begin{tikzpicture}
		\begin{axis}[
		font=\scriptsize,
		width=1.1\linewidth,
		height=2in,
		xlabel = $\delta u^{\text{max}}$,
		ylabel = {$\%$ erroneous pixels},
		xmin=0.1, xmax=1.0,
		ymin=0, ymax=45,
		legend pos=south east,
		legend style={nodes={scale=0.6, transform shape}},
		cycle list name=color
		]
		\addplot[color=cyan, mark=x]
		coordinates {
			(0.10,42.63)(0.20,37.80)(0.30,35.88)(0.40,35.32)(0.50,35.50)(0.60,35.52)(0.70,35.77)(0.80,36.00)(0.90,36.36)(1.00,36.77)};
		\addlegendentry{N=2}
		\addplot[color=blue, mark=asterisk]
		coordinates {
			(0.10,28.86)(0.20,26.51)(0.30,26.38)(0.40,26.33)(0.50,26.56)(0.60,26.99)(0.70,27.60)(0.80,28.42)(0.90,29.10)(1.00,30.07)};
		\addlegendentry{N=5}
		\addplot[color=red, mark=triangle]
		coordinates {
			(0.10,18.37)(0.20,17.24)(0.30,17.23)(0.40,17.65)(0.50,18.50)(0.60,19.95)(0.70,22.21)(0.80,23.80)(0.90,24.33)(1.00,24.96)};
		\addlegendentry{N=10}
		\addplot[color=green, mark=+]
		coordinates {
			(0.10,5.51)(0.20,7.09)(0.30,11.56)(0.40,14.94)(0.50,16.07)(0.60,18.95)(0.70,20.76)(0.80,22.59)(0.90,23.09)(1.00,23.90)};
		\addlegendentry{N=50}
		\addplot[color=magenta, mark=o]
		coordinates {
			(0.10,5.66)(0.20,11.71)(0.30,16.37)(0.40,19.45)(0.50,23.24)(0.60,25.98)(0.70,27.02)(0.80,28.00)(0.90,28.46)(1.00,28.67)};
		\addlegendentry{N=100}
		%\legend{N=2}
		\end{axis}
		\end{tikzpicture}%}
		%\fbox{\rule{0pt}{1.5in} \rule{0.9\linewidth}{0pt}}
		%		\includegraphics[width=1.0\linewidth]{images/plotlimitrange2.png}
		\caption{Accuracy of disparity (percentage of erroneous pixels, $\tau$) with limiting the magnitude of $\delta u$ for different warping iteration values $N$.}
		\label{fig:plotlimitingrange}
	\end{minipage}
	%\caption{Limiting the magnitude of $\delta u$ per iteration reduces the error around sharp image gradients and occlusion boundaries.}
	%{fig:limitingrange}
\end{figure*}

\newcommand{\thiswidth}{0.13}
\setlength\tabcolsep{2pt}%
\begin{figure}
	\scriptsize
	\centering
	\begin{tabular}{cccll}
		Ours ($180^{\circ}$) & $90^{\circ}$ & $165^{\circ}$ &&\\
		\includegraphics[width=\thiswidth\textwidth]{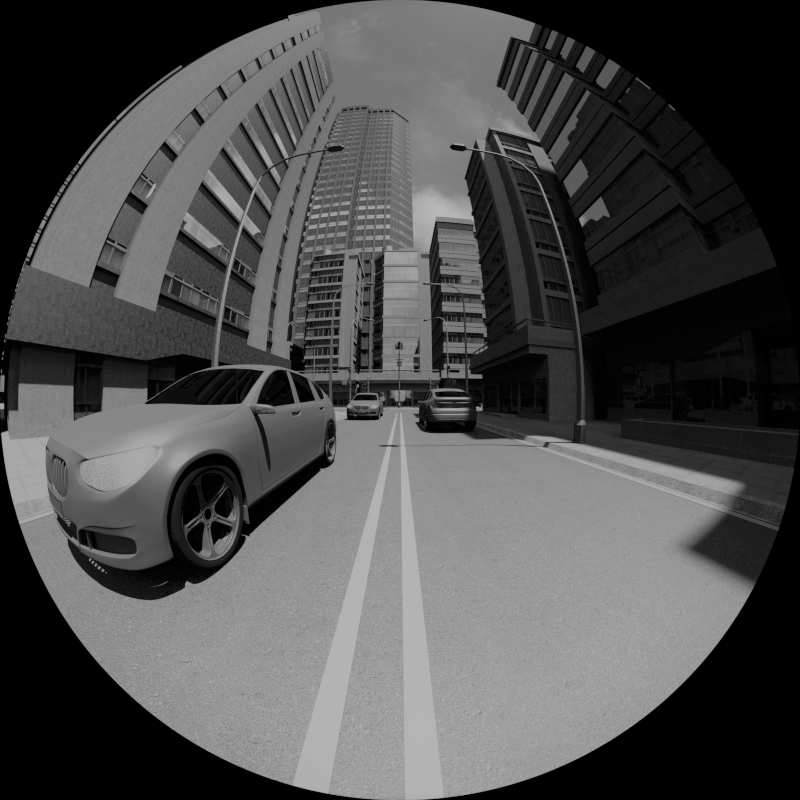} &
		\includegraphics[width=\thiswidth\textwidth]{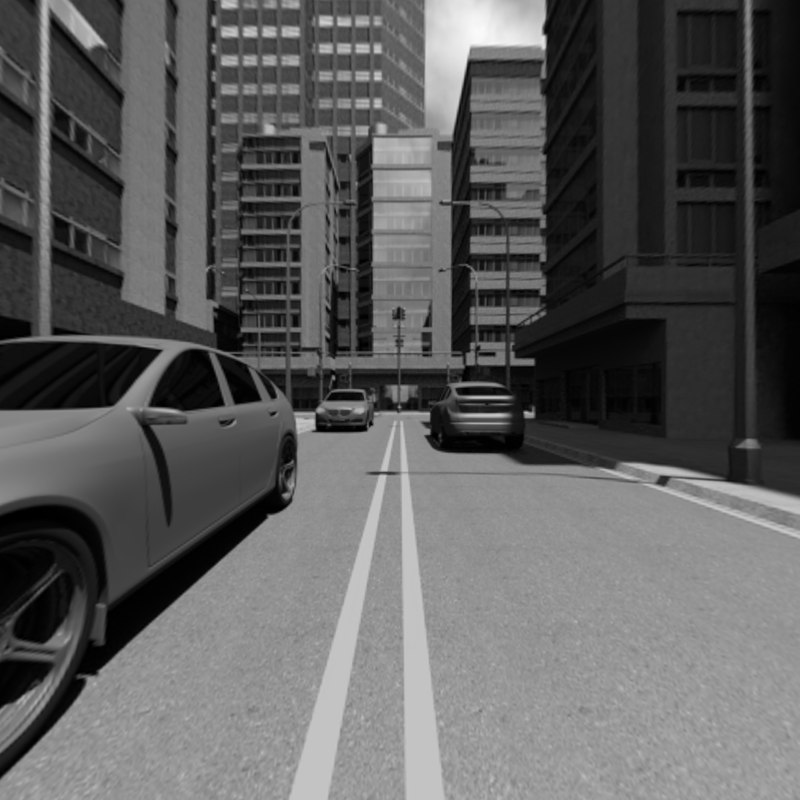} &
		\includegraphics[width=\thiswidth\textwidth]{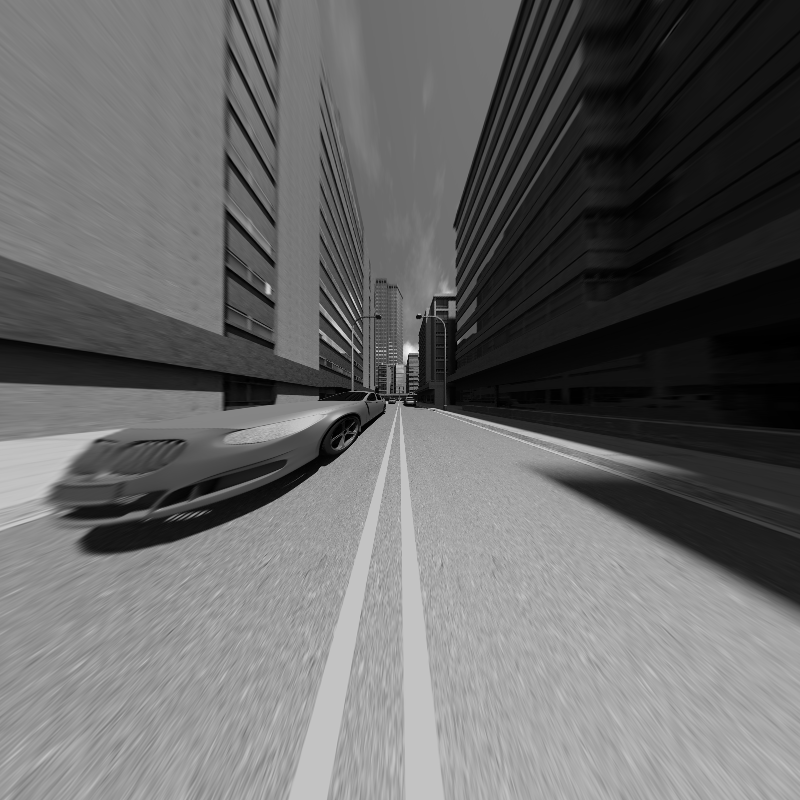} &&\\
		
		\begin{tikzpicture}
		\draw (0, 0) node[inner sep=0] {
			\includegraphics[width=\thiswidth\textwidth]{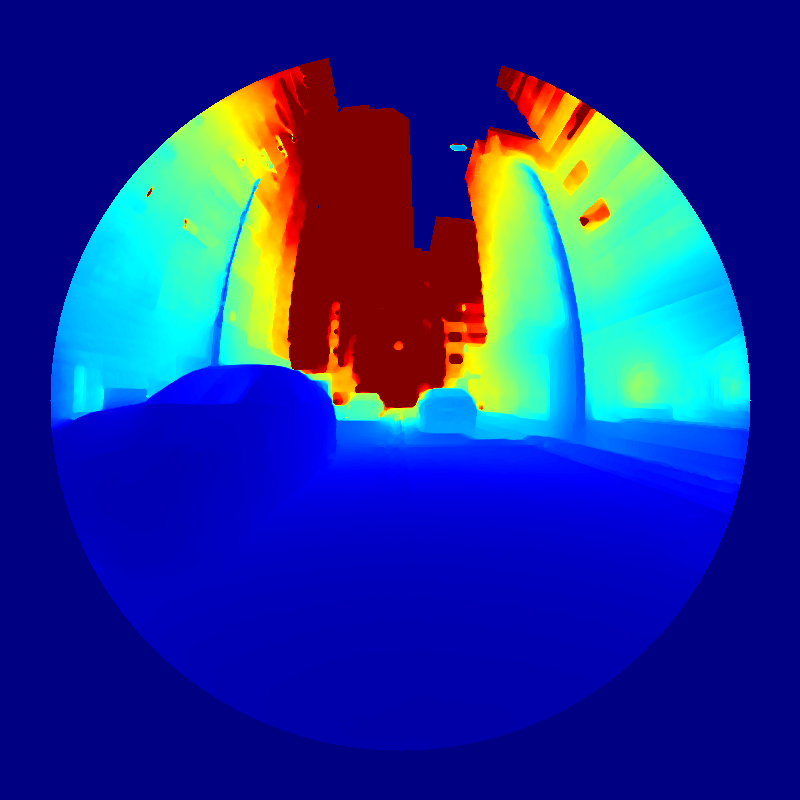}};
		\draw (-0.8, 1) node {\textcolor{white}{Depth}}; 
		\end{tikzpicture}&
		\includegraphics[width=\thiswidth\textwidth]{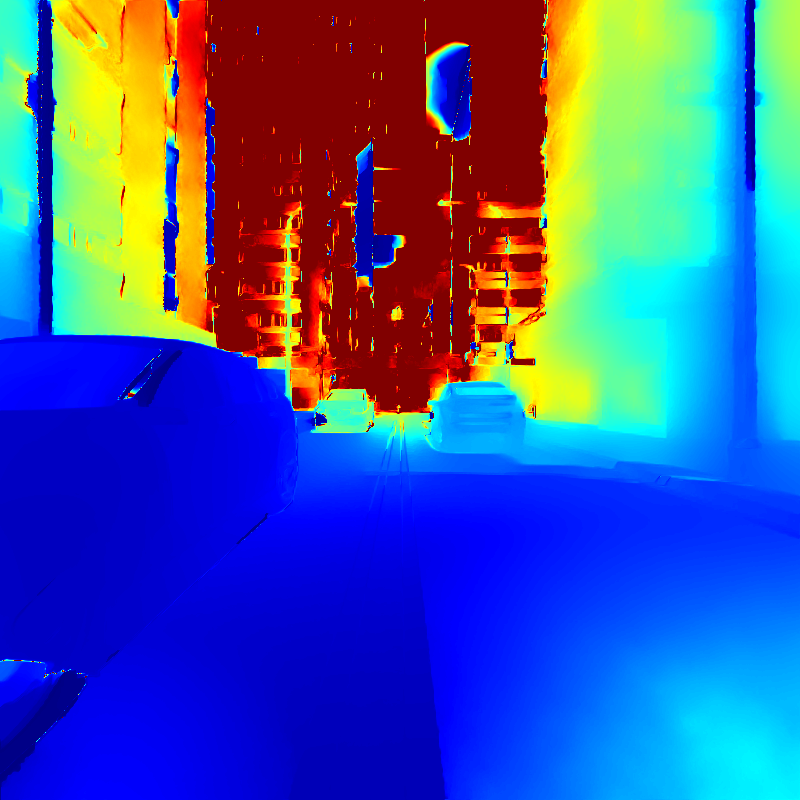} &
		\includegraphics[width=\thiswidth\textwidth]{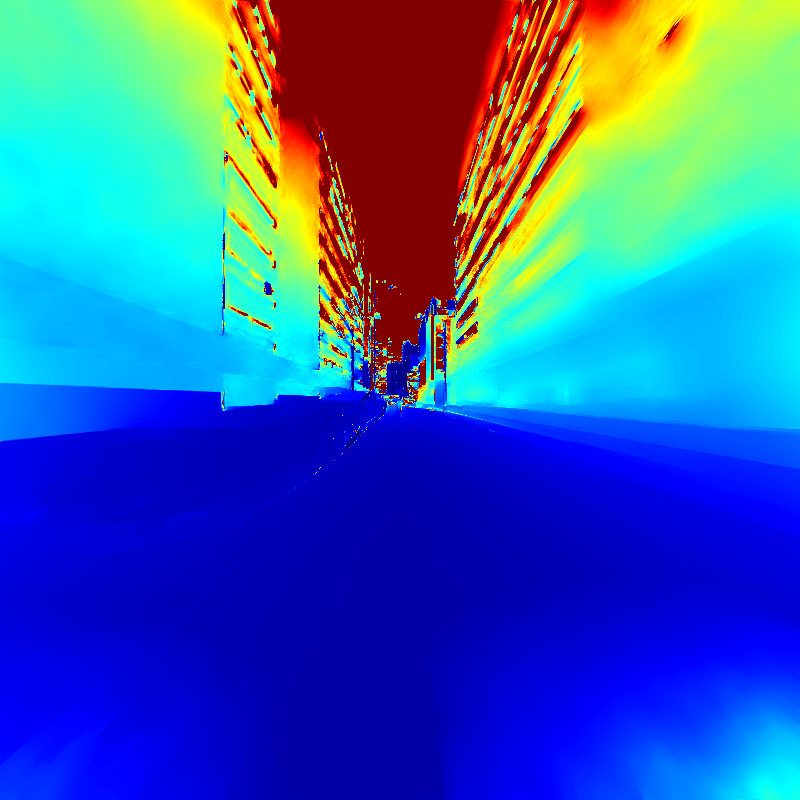} &&\\
		
		\multirow{7}{*}{
			\begin{tikzpicture}
			\draw (0, 0) node[inner sep=0] {
				\includegraphics[width=\thiswidth\textwidth]{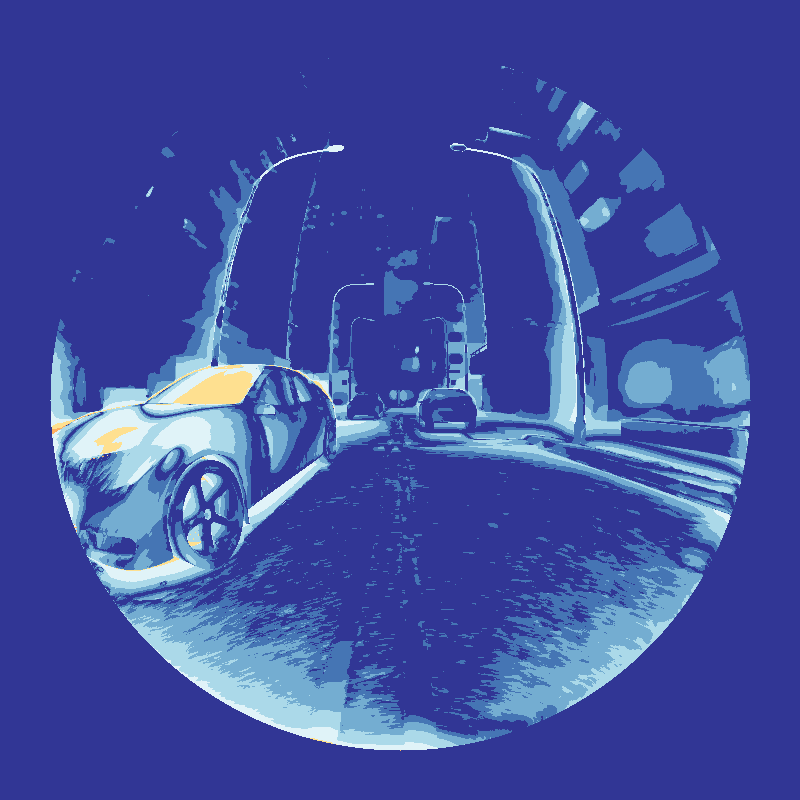}};
			\draw (-0.8, 1) node {\textcolor{white}{Error}}; 
			\end{tikzpicture}}&
		
		\multirow{7}{*}{
			\begin{tikzpicture}
			\draw (0, 0) node[inner sep=0] {
				\includegraphics[width=\thiswidth\textwidth]{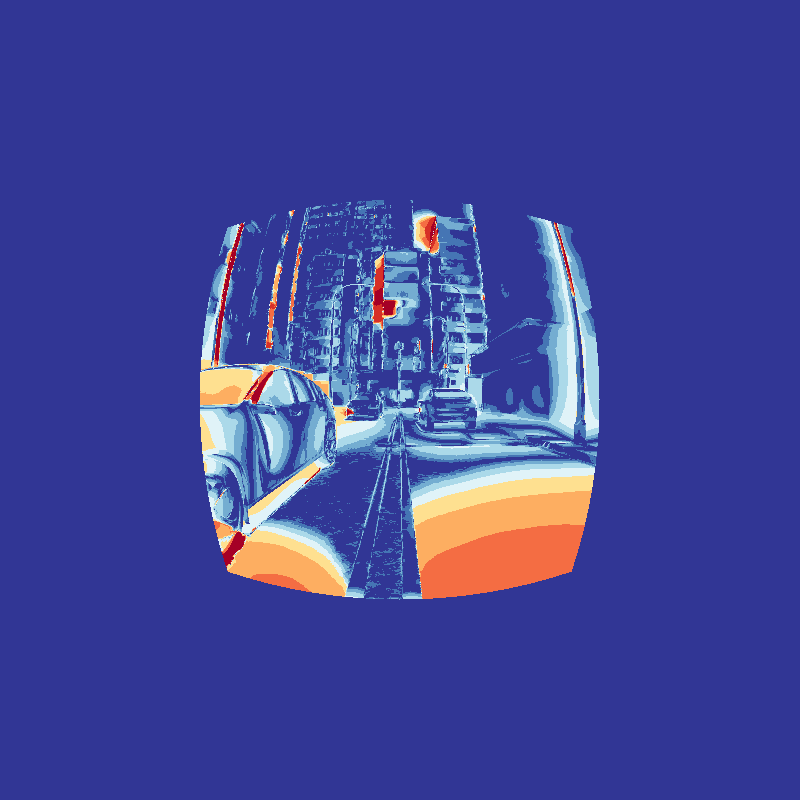}};
			\draw (0.65, -1) node {\textcolor{white}{+62.33\%}}; 
			\end{tikzpicture}}&
		
		\multirow{7}{*}{
			\begin{tikzpicture}
			\draw (0, 0) node[inner sep=0] {
				\includegraphics[width=\thiswidth\textwidth]{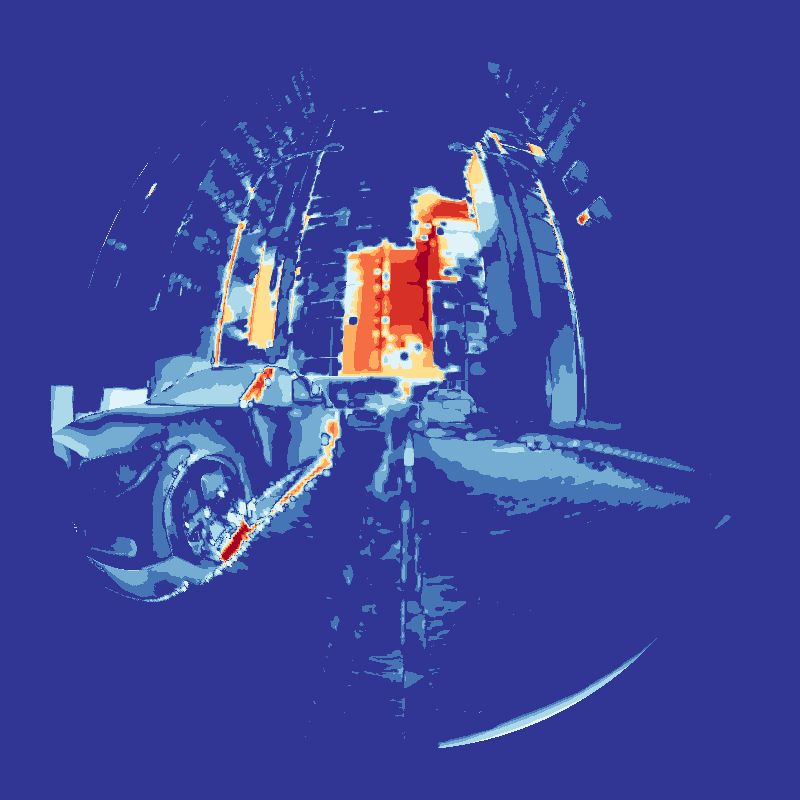}};
			\draw (0.65, -1) node {\textcolor{white}{+7.75\%}}; 
			\end{tikzpicture}} &
		
		\multirow{7}{*}{\includegraphics[width=0.008\textwidth]{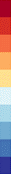}} &
		{\tiny $\infty$} \\	&&&&{\tiny 48} \\[2pt]  &&&&{\tiny 24} \\[2pt]  &&&&{\tiny 3} \\[2pt]  &&&&{\tiny 0.75} \\[2pt]  &&&&{\tiny 0.19} \\   &&&&{\tiny 0}\\[3pt]
	\end{tabular}
	\caption{Comparison between \cite{ranftl2012} with different field-of-view ($90^{\circ}$ and $165^{\circ}$) and our method. We compare the compare the disparity error \cite{kitti} as well as percentage of accuracy improvement by using our method.}
	\label{fig:additionalpixels}
\end{figure}
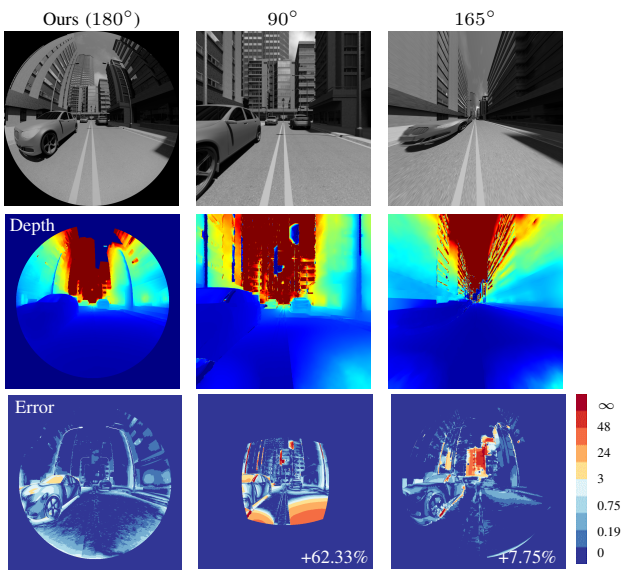

\section{RESULTS}
We present our results in the following sections. First, we show the effect of limiting the magnitude of the incremental disparity solution per warping iteration to the accuracy of the estmation. Then, we compare our method with an existing rectified variational stereo method and a discrete stereo matching method using both synthetic and real datasets with ground truth depth. Finally, we show some sample qualitative results on a commercial-off-the-shelf stereo camera fisheye system.

\subsection{Limiting Incremental Disparity}
To test the effect of limiting the incremental disparity, we measure the accuracy of our method on varying warping iteration and disparity limits. In Figure \ref{fig:limitingrange}, we show the photometric error (absolute difference between $I_0$ and warped $I_1$) when $\delta u^{max}=1.0px$ and $\delta u^{max}=0.2px$. From the images, we can see that the photometric error is larger in areas with significant information (e.g. intensity edges and occlusion boundaries) when $\delta u^{max}=1.0px$ compared to $\delta u^{max}=0.2px$. This happens because it is faster for the optimization to converge in highly textured surfaces which results in overshooting from the tracked epipolar curve, as shown in Figure \ref{fig:piecewiselinear}.

However, limiting the magnitude of $\delta u$ has an obvious drawback. If the warping iteration is not sufficient, the estimated $\delta u$ will not reach to its correct value which will result in higher error. We show this effect in Figure \ref{fig:plotlimitingrange}. Here, we plot various warping iterations $N$ and show the accuracy of estimation with increasing $\delta u^{max}$ using percentage of erroneous pixel measure ($\tau>1$) \cite{kitti}. Clearly, higher $N$ and smaller $\delta u^{max}$ results in a more accurate estimation.

%\begin{figure}
%	\begin{center}
%		%\fbox{\rule{0pt}{1.5in} \rule{0.9\linewidth}{0pt}}
%		\includegraphics[width=0.5\linewidth]{images/limitrangeplot.png}
%	\end{center}
%	\caption{Limiting the magnitude of $\delta u$ per iteration reduces the error around sharp image gradients and occlusion boundaries.}
%	\label{fig:limitingrangeplot}
%\end{figure}
\begin{figure*}
	\scriptsize
	\centering
	\begin{tabular}{ccccccll}
		Input & GT & Depth & Error & Depth & Error &&\\
		\includegraphics[width=0.15\textwidth]{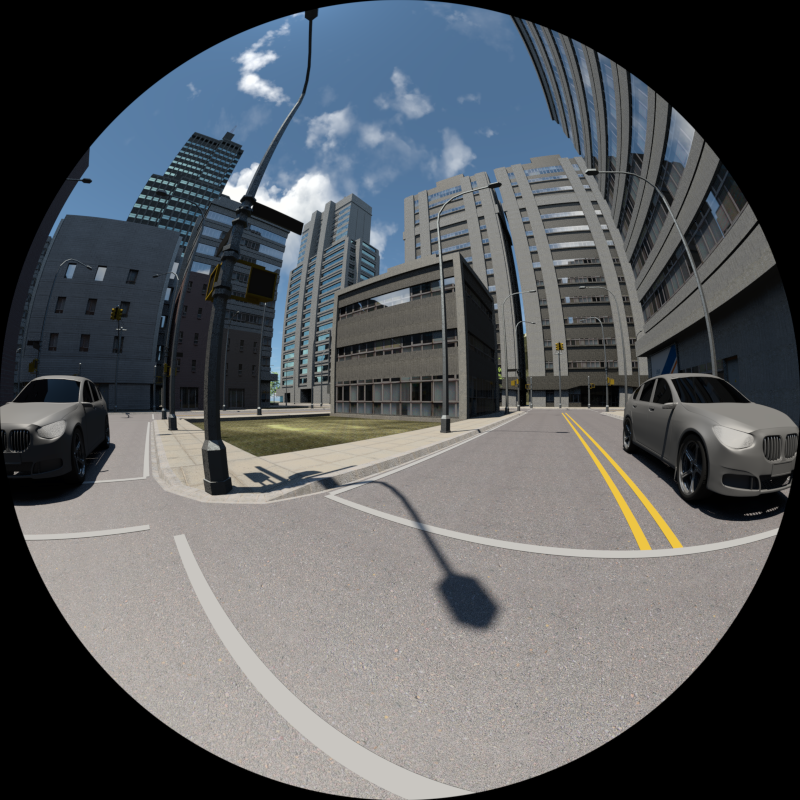} &
		\includegraphics[width=0.15\textwidth]{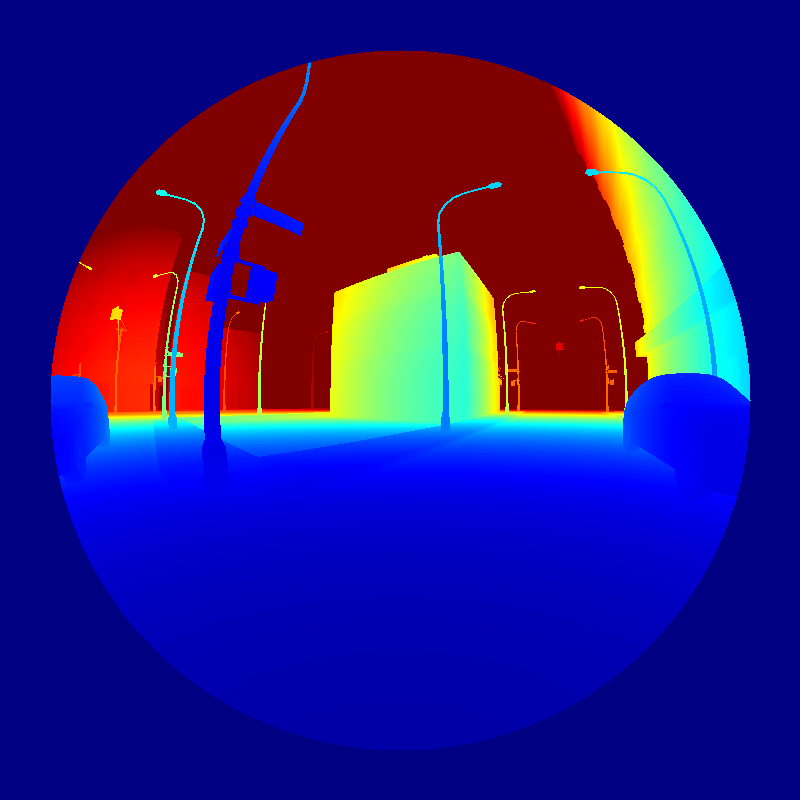} &
		\includegraphics[width=0.15\textwidth]{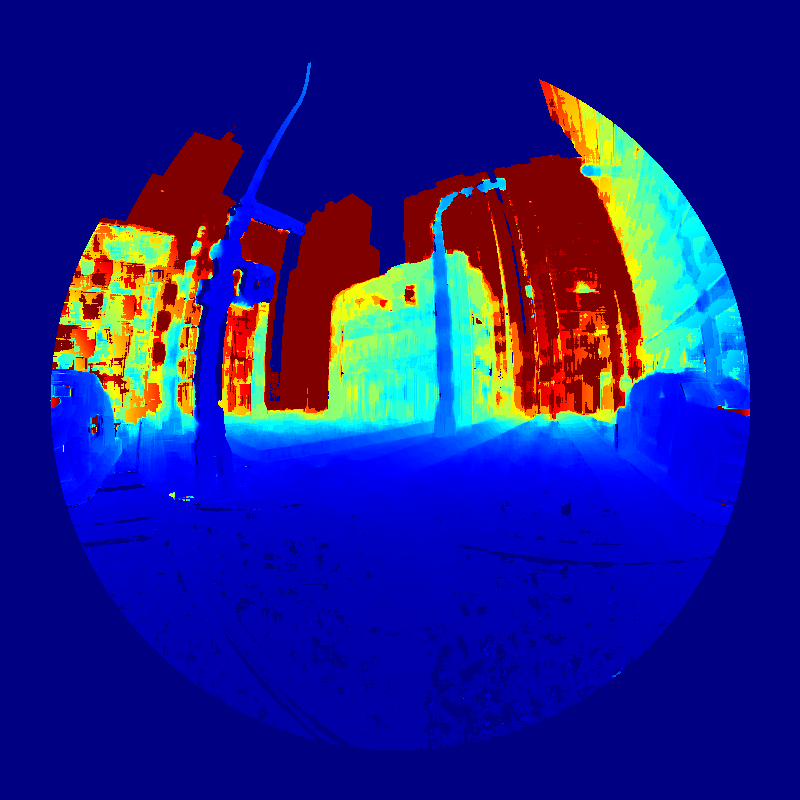} &
		\includegraphics[width=0.15\textwidth]{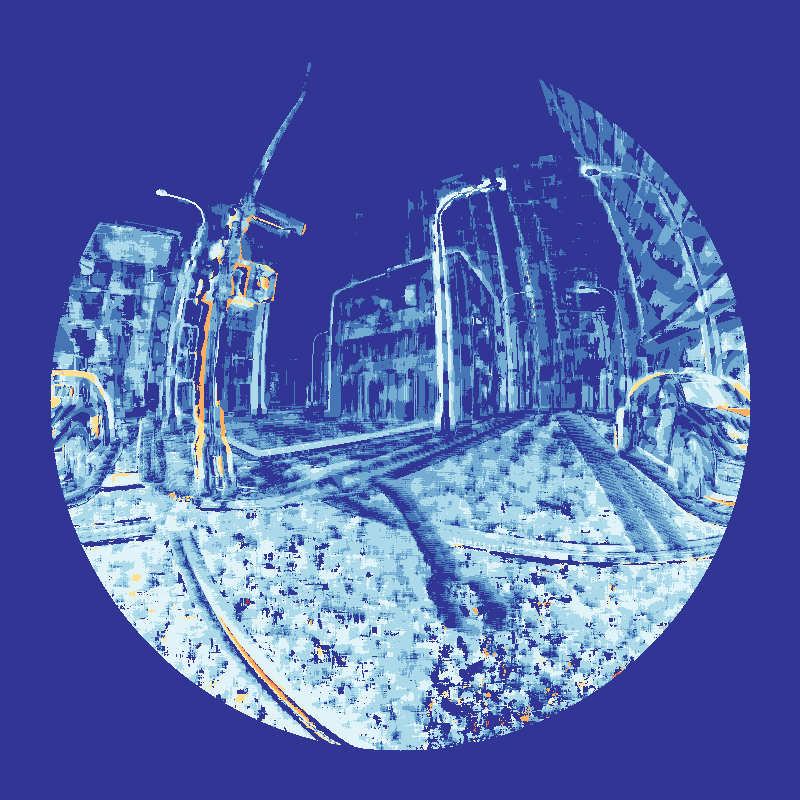}&
		\includegraphics[width=0.15\textwidth]{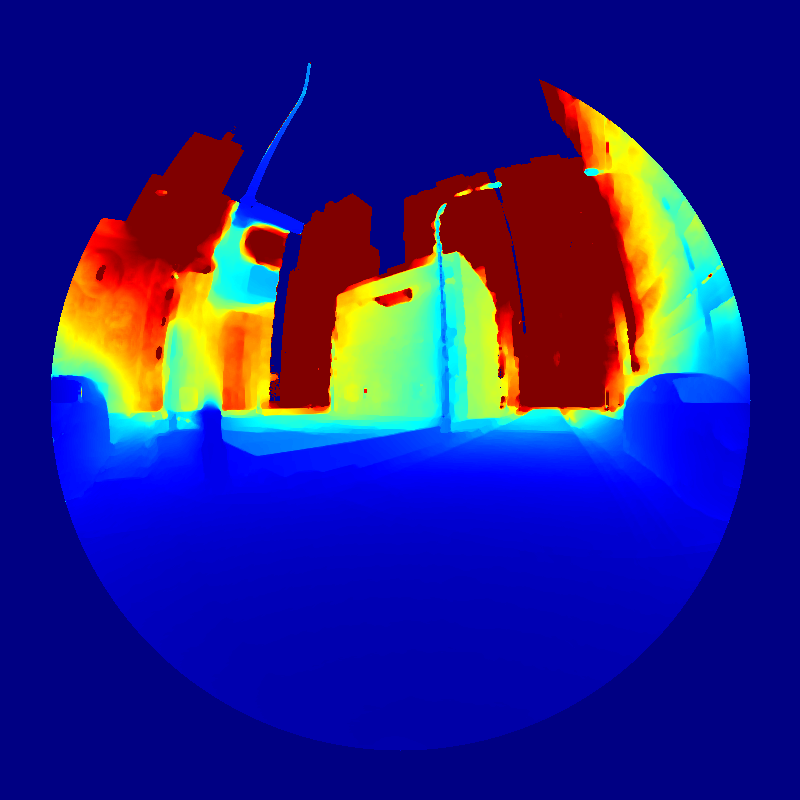} &
		\includegraphics[width=0.15\textwidth]{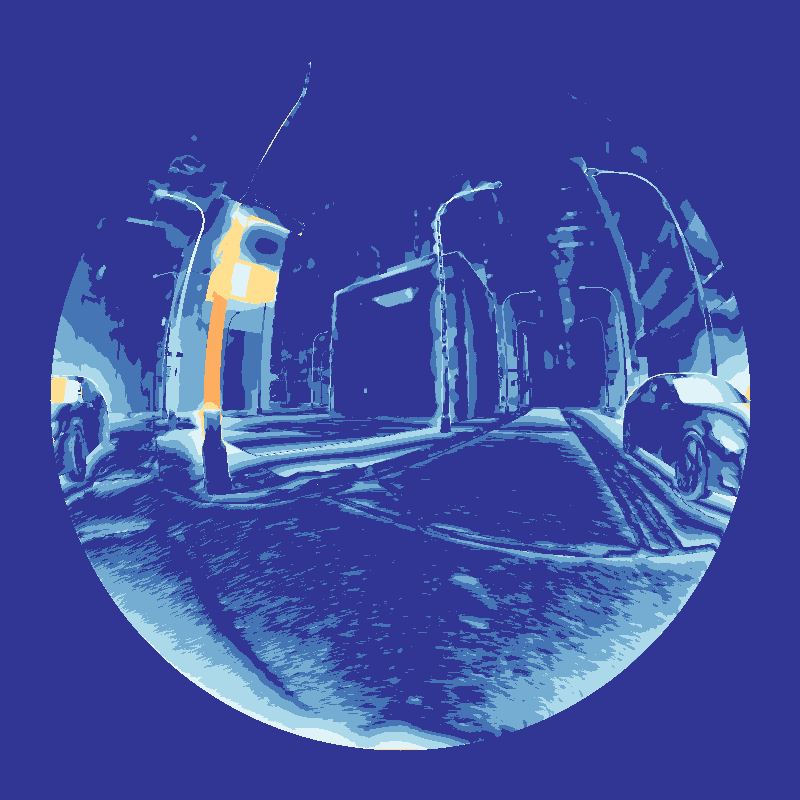} && \\

		\multirow{7}{*}{\includegraphics[width=0.15\textwidth]{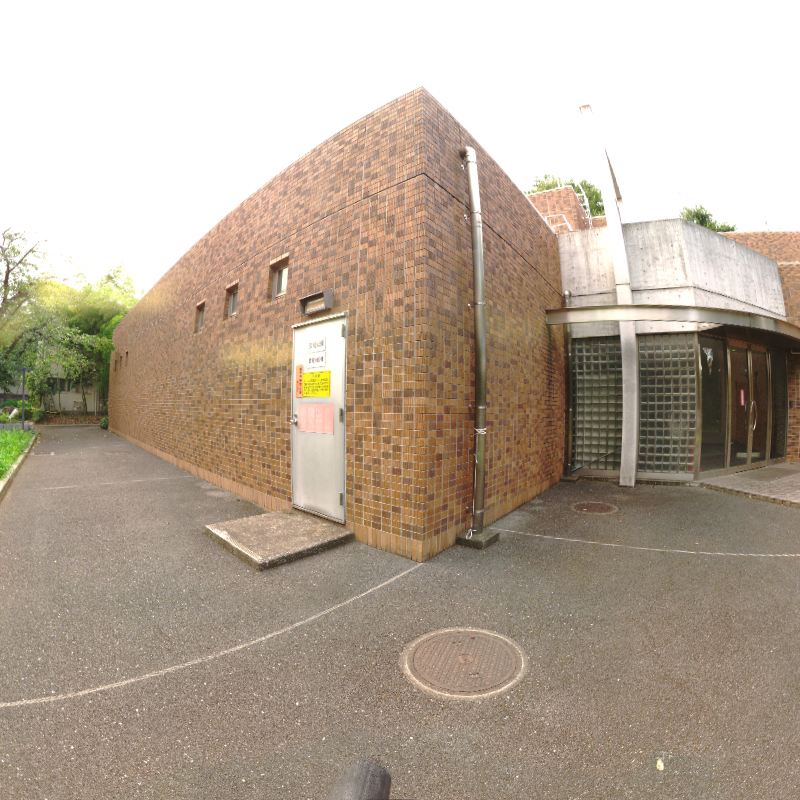}} &
		\multirow{7}{*}{\includegraphics[width=0.15\textwidth]{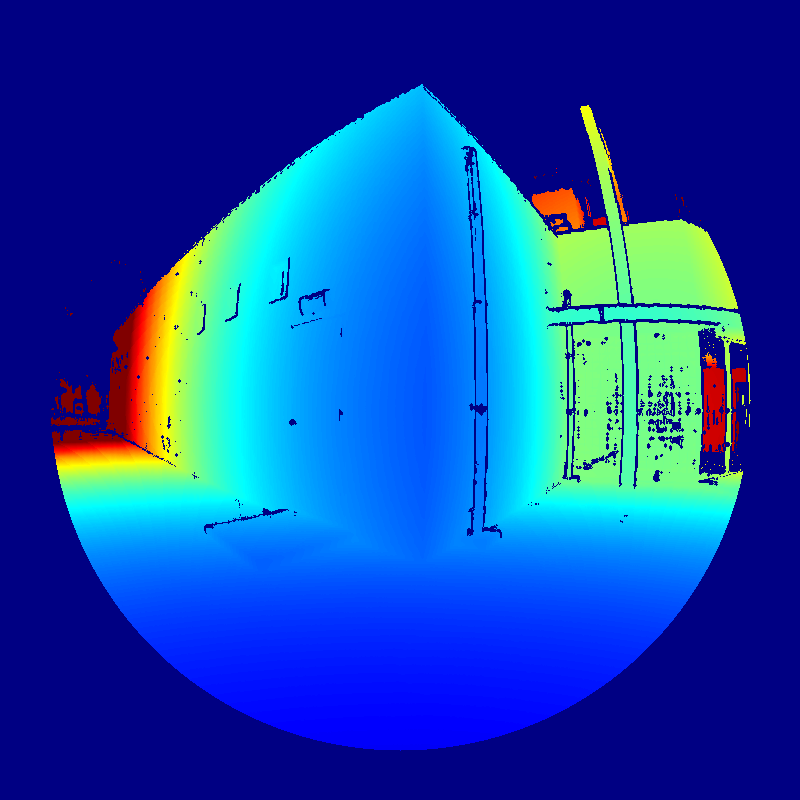}} &
		\multirow{7}{*}{\includegraphics[width=0.15\textwidth]{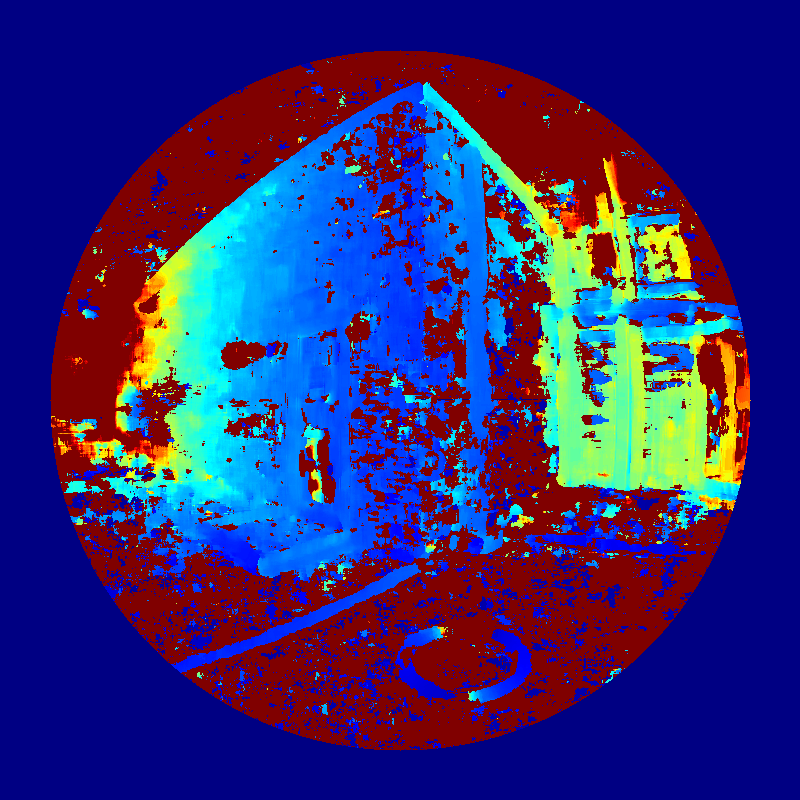}} &
		\multirow{7}{*}{\includegraphics[width=0.15\textwidth]{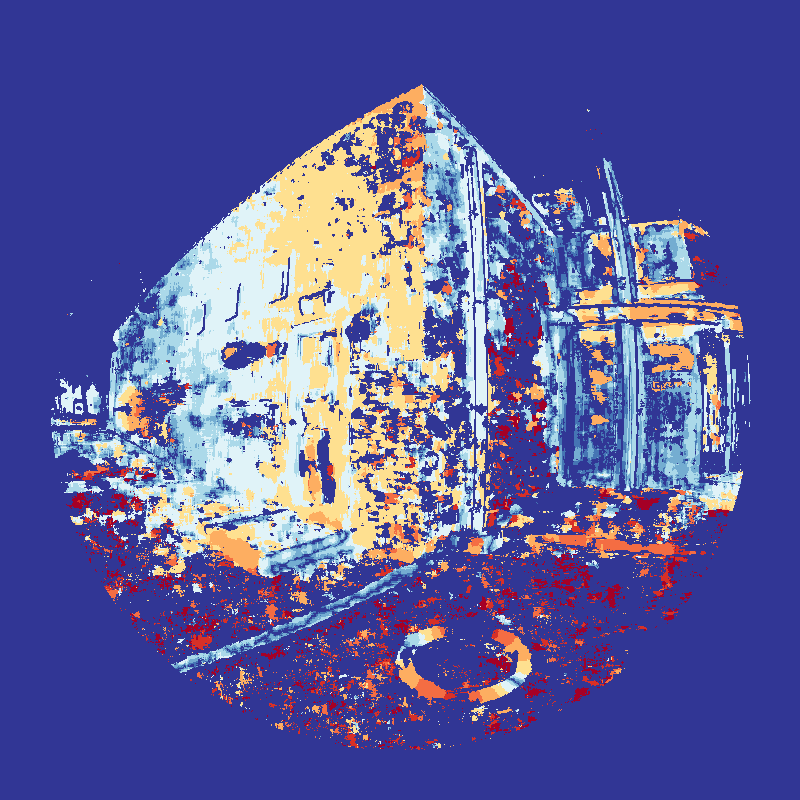}} &
		\multirow{7}{*}{\includegraphics[width=0.15\textwidth]{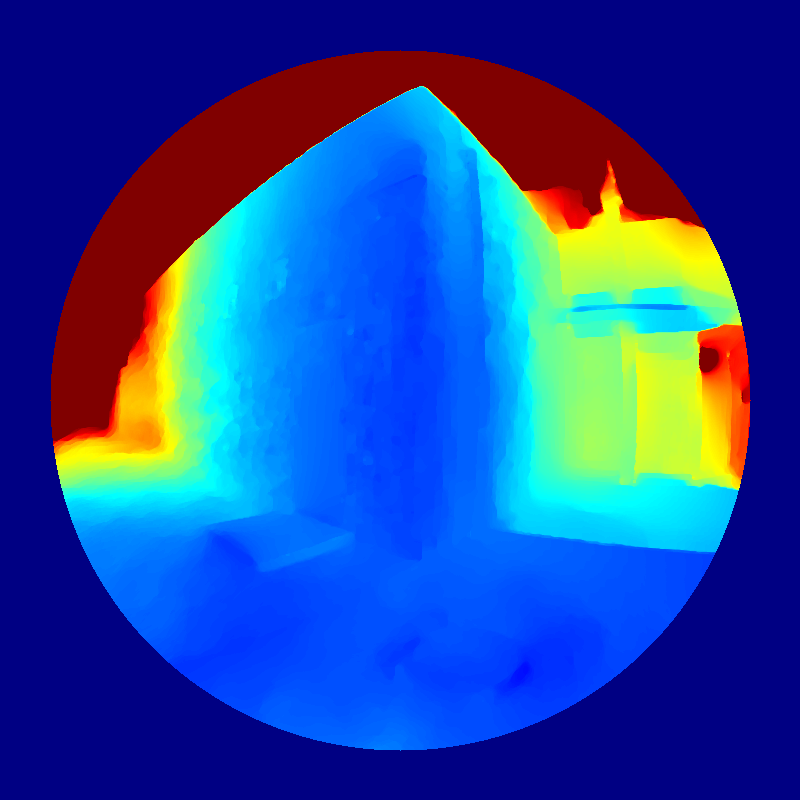}} &
		\multirow{7}{*}{\includegraphics[width=0.15\textwidth]{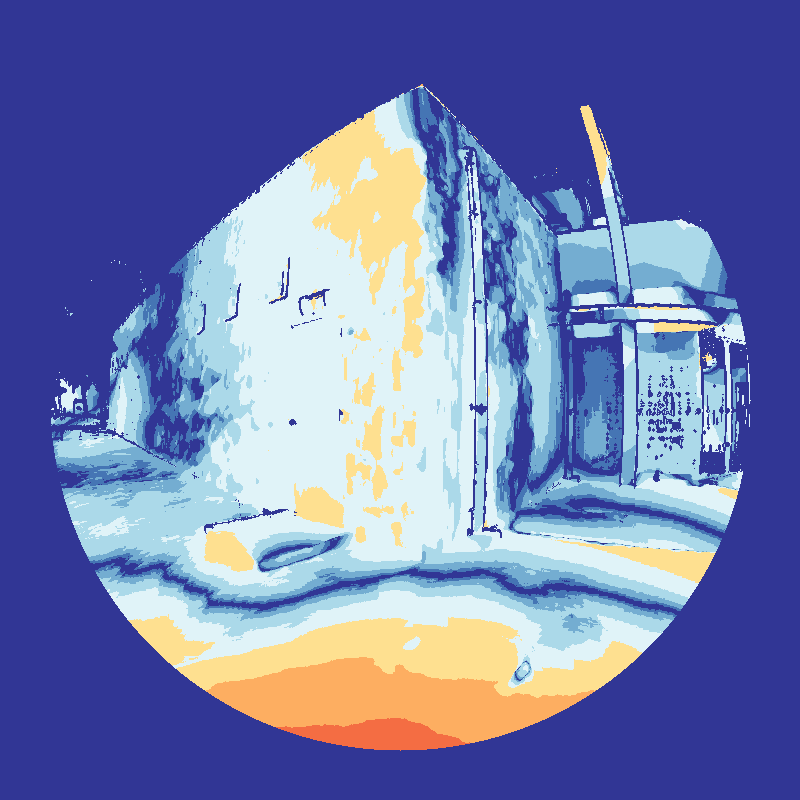}} &
		\multirow{7}{*}{\includegraphics[width=0.0095\textwidth]{images/legend.png}} &
		{\tiny $\infty$} \\	&&&&&&&{\tiny 48} \\[5pt]  &&&&&&&{\tiny 24} \\[5pt]  &&&&&&&{\tiny 3} \\[5pt]  &&&&&&&{\tiny 0.75} \\[5pt]  &&&&&&&{\tiny 0.19} \\   &&&&&&&{\tiny 0}\\[3pt]
		%	\begin{tabular}{l}
		%		$\infty$ \\	48 \\ 24 \\ 3 \\ 0.75 \\ 0.19 \\  0\\
		%%		\multirow{2}{*}{\includegraphics[width=0.01\textwidth]{images/legend.png}} & 
		%	\end{tabular}\\
		
		&& \multicolumn{2}{c}{\cite{fisheyeplanesweep2014}} & \multicolumn{2}{c}{Ours} &&
	\end{tabular}
	\caption{Sample results on real and synthetic data with \cite{fisheyeplanesweep2014} and our method with disparity error \cite{kitti}.}
	\label{fig:comparisonwithplanesweep}
\end{figure*}

\subsection{Comparison with Rectified Method}
We first compare our proposed approach with a rectified stereo method. To achieve a fair comparison, we use the same energy function and parameters in our implementation, except that we apply them in a rectified image. This rectified stereo approach is similar to the method presented in \cite{ranftl2012}, except that we use intensity values instead of the census transform. We also explicitly applied a time-step pre-conditioning step and a relaxation after every iteration.

We compare our method with varying FOV for \cite{ranftl2012} on a synthetic dataset \cite{sweepnet}. We use the same erroneous pixel measure from the previous section and summarize the result in Table \ref{tab:comparisonwithplanesweep} using an FOV of $120^{\circ}$ for \cite{ranftl2012}. We also compare the disparity error \cite{kitti} as well as the improvement additional accurate pixels (see Figure \ref{fig:additionalpixels}) using the full $180^{\circ}$ for our method and a FOV of $90^{\circ}$ and $165^{\circ}$ for \cite{ranftl2012}. 
%We summarize the results in Table \ref{tab:comparisonwithplanesweep} and Figure \ref{fig:additionalpixels}. We used the full $180^{\circ}$ for our method and compared it with $90^{\circ}$ and $165^{\circ}$ FOV for the rectified method. Both are run with the same number of iterations and image size (800x800). 

To better visualize the comparison, we transform the rectified error back to the original fisheye form. From the results, extreme compression around the center with ultra-wide angle ($165^{\circ}$) rectification results in higher error especially for distant objects. With larger image area coverage, our approach do not suffer from this compression problem and maintains uniform accuracy throughout the image. Moreover, with the lower compression around the center ($90^{\circ}$), the rectified method have increased error around the edges for closer objects (ground) due to increased displacement.

Additionally, we found no significant difference in processing time because the warping techniques are both run in a single GPU kernel call and consumes the same texture memory access latency.

\subsection{Comparison with Non-Rectified Method}
In this section, we compare our method with planesweep implemented on a fisheye camera system \cite{fisheyeplanesweep2014} on real and synthetic scenes. The images were captured from two arbitrary camera location with non-zero translation. We show the sample results in Figure \ref{fig:comparisonwithplanesweep} and Table \ref{tab:comparisonwithplanesweep}.

One of the advantages of variational methods is the inherent density of the estimated depth when compared to discrete matching methods. In our experiments, we found that while our method is denser and significantly smoother than \cite{fisheyeplanesweep2014}, it is more prone to miss very thin objects such as poles. Moreover, because our method is built upon a pyramid scheme, very large displacements are difficult to estimate which is visible in the results when the object is very close to the camera (nearest ground area).

Nevertheless, we show in Table \ref{tab:comparisonwithplanesweep} that our method is overall more accurate compared to \cite{fisheyeplanesweep2014} even after we removed the ambigous pixels due to occlusion and left-right inconsistency. (In Table \ref{tab:comparisonwithplanesweep}, we use only the valid pixels in \cite{fisheyeplanesweep2014} for comparison).

%\begin{figure*}
%	\begin{center}
%		%\fbox{\rule{0pt}{3.0in} \rule{0.9\linewidth}{0pt}}
%		\includegraphics[width=1.0\linewidth]{images/results3.png}
%	\end{center}
%	\caption{Results on synthetic and real dataset with \cite{fisheyeplanesweep2014} and our method.}
%	\label{fig:comparisonwithplanesweep}
%\end{figure*}

\subsection{Real-World Test}
We tested our method on a laptop computer with NVIDIA GTX1060 GPU and an Intel RealSense T265 stereo camera, which has a $163\pm5^{\circ}$ FOV, global-shutter 848x800 grayscale image and a 30fps throughput. We show the sample results in Figure \ref{fig:stereocamera}. We were able to achieve a 10fps with 5 warping iterations on a full image, and 30fps with 20 warping iterations on a half-size image. This system can be easily mounted on medium sized rover for SLAM applications.

\section{CONCLUSION AND FUTURE WORK}
In this paper, we presented a warping technique for handling fisheye cameras specified for real-time variational stereo estimation methods without explicit image rectification. From our results, we showed that our approach can achieve higher and more uniform accuracy and larger FOV compared to conventional methods without increasing the processing time.

Because of the wider FOV of fisheye cameras, the disadvantage of most variational methods, which is handling large displacement (wide baseline or near objects), is highlighted. However, this can be overcome by using large displacement techniques or initialization with discreet methods (such as planesweep).

\setlength\tabcolsep{1pt}%
\begin{table}
	\caption{Disparity Error Comparison with Rectified, Non-rectified, and our method.}
	\label{tab:comparisonwithplanesweep}
	\begin{center}
		\begin{tabular}{cc|cc|cc|cc} 
			\toprule
			&Frame &   \multicolumn{2}{c}{\cite{ranftl2012}} & \multicolumn{2}{c}{\cite{fisheyeplanesweep2014}} &\multicolumn{2}{c}{Ours}\\
			\midrule
			&& $\tau > 1$   & $\tau > 3$    & $\tau > 1$   & $\tau > 3$ & $\tau > 1$   & $\tau > 3$\\
			\multirow{4}{*}{synthetic}&04 & 14.72 & 8.06 & 28.46 & 3.40 & \textbf{6.60} & \textbf{0.78}\\
			&05 & 17.19 & 10.06 & 26.13 & 2.05 & \textbf{7.66} & \textbf{0.97}\\
			&06 & 11.76 & 8.33  & 27.76 & 1.55 &\textbf{5.51} & \textbf{1.51} \\
			&07 & 11.21 & 2.46  & 27.46 & 1.75 &\textbf{4.55} & \textbf{0.28}\\
			\midrule
			\multirow{2}{*}{real}&92   &  - &  -  & -&50.64&  - & \textbf{34.79} \\
			&100   &  - &  -  & -  &43.85&-& \textbf{20.15} \\
			\bottomrule
		\end{tabular}
	\end{center}
\end{table}


%%%%%%%%%%%%%%%%%%%%%%%%%%%%%%%%%%%%%%%%%%%%%%%%%%%%%%%%%%%%%%%%%%%%%%%%%%%%%%%%

%%%%%%%%%%%%%%%%%%%%%%%%%%%%%%%%%%%%%%%%%%%%%%%%%%%%%%%%%%%%%%%%%%%%%%%%%%%%%%%%

%%%%%%%%%%%%%%%%%%%%%%%%%%%%%%%%%%%%%%%%%%%%%%%%%%%%%%%%%%%%%%%%%%%%%%%%%%%%%%%%

%%%%%%%%%%%%%%%%%%%%%%%%%%%%%%%%%%%%%%%%%%%%%%%%%%%%%%%%%%%%%%%%%%%%%%%%%%%%%%%%

\bibliographystyle{IEEEtran}
\bibliography{IEEEabrv,ref}

\begin{thebibliography}{10}
\providecommand{\url}[1]{#1}
\csname url@rmstyle\endcsname
\providecommand{\newblock}{\relax}
\providecommand{\bibinfo}[2]{#2}
\providecommand\BIBentrySTDinterwordspacing{\spaceskip=0pt\relax}
\providecommand\BIBentryALTinterwordstretchfactor{4}
\providecommand\BIBentryALTinterwordspacing{\spaceskip=\fontdimen2\font plus
\BIBentryALTinterwordstretchfactor\fontdimen3\font minus
  \fontdimen4\font\relax}
\providecommand\BIBforeignlanguage[2]{{%
\expandafter\ifx\csname l@#1\endcsname\relax
\typeout{** WARNING: IEEEtran.bst: No hyphenation pattern has been}%
\typeout{** loaded for the language `#1'. Using the pattern for}%
\typeout{** the default language instead.}%
\else
\language=\csname l@#1\endcsname
\fi
#2}}

\bibitem{stuhmer2010}
J.~St{\"u}hmer, S.~Gumhold, and D.~Cremers, ``Real-time dense geometry from a
  handheld camera,'' \emph{Pattern Recognition. DAGM 2010. LNCS.}, vol. 6376,
  2010.

\bibitem{ranftl2012}
R.~Ranftl, S.~Gehrig, T.~Pock, and H.~Bischof, ``Pushing the limits of stereo
  using variational stereo estimation,'' in \emph{Proc. {IEEE} Intel. Vehic.},
  June 2012.

\bibitem{schneider2016}
J.~Schneider, C.~Stachniss, and W.~Forstner, ``On the accuracy of dense fisheye
  stereo,'' \emph{IEEE Robotics and Automation Letters}, vol.~1, no.~1, 2016.

\bibitem{arican2007}
Z.~Arican and P.~Frossard, ``Dense disparity estimation from omnidirectinoal
  images,'' in \emph{IEEE Conf. Adv. Vid. Sig. Based Surv.}, September 2007.

\bibitem{schonbein2014}
M.~Schonbein and A.~Geiger, ``Omnidirectional 3d reconstruction in augmented
  manhattan worlds,'' in \emph{Proc. {IEEE} Int. Work. Robots Sys.}, 2014.

\bibitem{bagnato2011}
L.~Bagnato, P.~Frossard, and P.~Vandergheynst, ``A variational framework for
  structure from motion in omnidirectional image sequences,'' \emph{J. Math
  Imaging Vis.}, vol.~41, pp. 182--193, 201.

\bibitem{gao2017}
W.~Gao and S.~Shen, ``Dual-fisheye omnidirectinoal stereo,'' in \emph{Proc.
  {IEEE} Int. Work. Robots Sys.}, 2017.

\bibitem{lsdslam2015}
D.~Caruso, J.~Engel, and D.~Cremers, ``Large-scale direct slam for
  omnidirectional cameras,'' in \emph{Proc. {IEEE} Int. Work. Robots Sys.},
  2015.

\bibitem{unifiedcameramodel}
C.~Geyer and K.~Daniilidis, ``A unifying theory for central panoramic systems
  and practical applications,'' in \emph{Proc. {IEEE} Europ. Conf. Comput.
  Vis.}, July 2000, pp. 445--461.

\bibitem{bunschoten2003}
R.~Bunschoten and B.~Krose, ``Robust scene reconstruction from an
  omnidirectinoal vision system,'' \emph{{IEEE} Trans. Robot. Automat.}, 2003.

\bibitem{khomutenko2016}
B.~Khomutenko, G.~Garcia, and P.~Martinet, ``Direct fisheye stereo
  correspondence using enhanced unified camera model and semi-global matching
  algorithm,'' in \emph{ICARCV}, 2016.

\bibitem{fisheyeplanesweep2014}
C.~Hane, L.~Heng, G.~H. Lee, A.~Sizov, and M.~Pollefeys, ``Real-time direct
  dense matching on fisheye images using plane-sweeping stereo,'' in
  \emph{Proc. Int. Conf. 3D Vis.}, December 2014.

\bibitem{inso2016}
S.~Im, H.~Ha, F.~Rameau, H.-G. Jeon, G.~Choe, and I.~S. Kweon, ``All-around
  depth from small motion with a spherical panoramic camera,'' in \emph{Proc.
  {IEEE} Europ. Conf. Comput. Vis.}, 2016, pp. 156--172.

\bibitem{sfmfisheye}
B.~Micusik and T.~Pajdla, ``Structure from motion with wide circular
  field-of-view cameras,'' \emph{{IEEE} Trans. Pattern Analysis and Machine
  Intelligence}, vol.~28, no.~7, pp. 1135--1149, July 2006.

\bibitem{censustransform}
R.~Zabih and J.~Li, ``Non-parametric local transforms for computing visual
  correspondence,'' in \emph{Proc. {IEEE} Europ. Conf. Comput. Vis.}, 1994, pp.
  151--158.

\bibitem{warping}
N.~Papenberg, A.~Bruhn, T.~Brox, S.~Didas, and J.~Weickert, ``Highly accurate
  optical flow computation with theoretically justified warping,'' \emph{Int.
  J. Comput. Vis.}, vol.~67, pp. 141--158, 2006.

\bibitem{svoboda1998}
T.~Svoboda, T.~Pajdla, and V.~Hlavac, ``Epipolar geometry for panoramic
  cameras,'' in \emph{Proc. {IEEE} Europ. Conf. Comput. Vis.}, 1998, pp.
  218--231.

\bibitem{enhanced}
B.~Khomutenko, G.~Garcia, and P.~Martinet, ``An enhanced unified camera
  model,'' \emph{IEEE Robotics and Automation Letters}, vol.~1, no.~1, pp.
  137--144, January 2016.

\bibitem{kannala2006}
J.~Kannala and S.~S. Brandt, ``A generic camera model and calibration method
  for conventional, wide-angle, fisheye lenses,'' \emph{Proc. {IEEE} Int. Conf.
  Comput. Vis.}, vol.~28, no.~8, pp. 1335--1340, September 2006.

\bibitem{preconditioning}
T.~Pock and A.~Chambolle, ``Diagonal pre-conditioning for first order
  primal-dual algorithms in convex optimization,'' in \emph{Proc. {IEEE} Int.
  Conf. Comput. Vis.}, 2011.

\bibitem{firstorderapprox}
A.~Chambolle and T.~Pock, ``A first-orer primal-dual algorithm for convex
  problems with applications to imaging,'' \emph{Journal of Mathematical
  Imaging and Vision}, vol.~40, no.~1, pp. 120--145, May 2011.

\bibitem{kitti}
A.~Geiger, P.~Lenz, C.~Stiller, and R.~Urtasun, ``Vision meets robotics: The
  kitti dataset,'' \emph{Int. J. Robot. Res.}, 2013.

\bibitem{sweepnet}
Z.~Zhang, H.~Rebecq, C.~Forster, and D.~Scaramuzza, ``Benefit of large
  field-of-view cameras for visual odometry,'' in \emph{Proc. {IEEE} Int. Conf.
  Robot Automat.}, 2016.

\end{thebibliography}

\end{document}